\documentclass[%
 reprint,
 superscriptaddress,
 amsmath,amssymb,
 aps,
 pra,
 longbibliography,
 floatfix,
]{revtex4-2}

\usepackage{graphicx}
\graphicspath{{figures/}}
\usepackage{float}
\usepackage{placeins}
\usepackage{needspace}
\usepackage{booktabs}
\usepackage{array}
\usepackage{xcolor}
\usepackage{amsthm}
\usepackage{hyperref}
\hypersetup{colorlinks=true,linkcolor={blue!55!black},
  citecolor={blue!55!black},urlcolor={teal!70!black}}

\newcommand{\iou}{\mathrm{IoU}}
\DeclareMathOperator{\Exec}{Exec}

\begin{document}

\title{From Pixels to PCells: A Neurosymbolic Approach to Photonic Component Creation}

\author{Aadarsh Agarwal}
\email[Correspondence to: ]{aadarwal@mit.edu}
\affiliation{Research Laboratory of Electronics, Massachusetts Institute of Technology, Cambridge, Massachusetts 02139, USA}
\affiliation{The College, University of Chicago, Chicago, Illinois 60637, USA}

\author{Kenaish Al Qubaisi}
\affiliation{Research Laboratory of Electronics, Massachusetts Institute of Technology, Cambridge, Massachusetts 02139, USA}

\author{Dirk Englund}
\affiliation{Research Laboratory of Electronics, Massachusetts Institute of Technology, Cambridge, Massachusetts 02139, USA}

\date{July 29, 2026}

\begin{abstract}
We present PixCell, a neurosymbolic system in which multimodal agents convert
a visually presented photonic component into a parametric program over a small
domain-specific language (DSL) of geometric primitives. A system enabling
deterministic visual verification renders evaluation asymmetrically cheaper
than the generation attempt. While models using multi-seed sampling and
iterative revision reach a mean best-turn IoU of only $0.416$, multimodal
agents through PixCell's interface and verifier consistently exceed $0.9$ mean
IoU, with scores reaching $0.974$ and $0.955$ across eight component targets
while also satisfying source contracts. These results demonstrate that frontier
multimodal agents can reliably understand and render executable parametric
representations from visual targets. Using these live parameters, cross-stack
studies on an interferometer reconstruct primitive programs that satisfy an
$8.0$\,nm free spectral range target and the original footprint constraint on
modeled 220-nm SOI, 400-nm SiN, and 400-nm TFLN stacks. PixCell further carries
a paper-derived splitter from visual reconstruction through SOI full-wave
simulation, producing symmetric propagation and balanced outputs. Finally, the
same executable verifier supplies a training reward and dataset used to train a
Qwen3.6-35B-A3B model with LoRA and GRPO without supervised demonstrations. On
eight training-excluded paper figures, its mean champion IoU rises from
$0.422$ after eight initial attempts to $0.491$ after three verifier-guided
revision rounds. These results therefore establish a controlled framework for
measuring, retargeting, and improving visual-to-parametric photonic component
design.
\end{abstract}

\maketitle
\raggedbottom

\section{Introduction}\label{sec:intro}

\begin{figure*}[t]
\centering
\includegraphics[width=\linewidth]{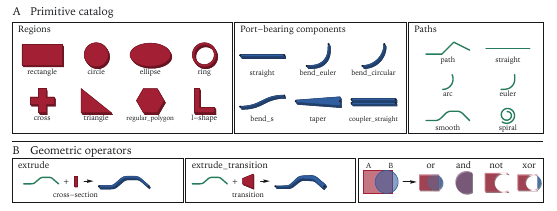}
\caption{\textbf{The PixCell geometric domain-specific language:
GDSFactory primitives and operators.}
(a)~The permitted catalog contains geometric regions, port-bearing
components, and one-dimensional paths, shown in red, blue, and green,
respectively.
(b)~Extrusion sweeps an independently editable cross-section along a path;
\texttt{extrude\_transition} varies that cross-section between its
endpoints. Boolean operations combine constructed regions through an
explicit set operation, with red and blue denoting operand-exclusive
regions and purple their intersection. The operator names match the
GDSFactory 9.20.7 interface used by the source gate, and every
function-relevant dimension must remain a named variable.}
\label{fig:primitives}
\end{figure*}

Foundation models have increasingly been combined with code generation and
tool use in the form of ``agents'' to solve complex and long-horizon science
and engineering tasks~\cite{react,sweagent,metr,alphaevolve}. The interface
for these problems has extended beyond text, with visual capabilities
enabling multimodal agents~\cite{gpt4,design2code,swebenchmm}. Such
multimodal intelligence has opened up new ways to approach science and
engineering tasks for problems that are fundamentally based on
visual-textual relationships. Photonic integrated circuit (PIC) design is a
particularly relevant problem of this nature.

PIC design involves specifications and netlists that describe component
connections. The behavior of each component is governed by wavelength-scale
geometry interpreted within a process design kit (PDK). These PDKs define
fabrication layers and design rules~\cite{chrostowski,bogaerts}. A reusable
component is therefore represented by layered geometry, ports, and
parameters for simulation, often expressed through parametric layout
code~\cite{gdsfactory}. For programmatic photonic IC design, this creates a
well-defined relationship between the visual structure of components and
the textual code used to construct them for larger circuits.

However, much of the work towards PIC design automation has focused on
language models operating through text-based systems. PICBench, for
instance, evaluates model-generated netlists for device- and circuit-level
problems~\cite{picbench}. PhIDO uses a multi-agent workflow to translate
natural-language requests into parametric netlists and GDSII
layouts~\cite{phido}. Even agent-based approaches involving
simulator-coupled agents that propose and revise designs against quantitative
performance criteria operate through code-based layout tools and accelerated
electromagnetic solvers~\cite{agentic,tidy3d}. These
systems support circuit assembly and optimization within named component
libraries or predefined parametric design spaces. Their evaluated interfaces
operate through textual specifications, netlists, or code-defined layouts
rather than component geometry supplied as a visual input. Yet the behavior
of a photonic component is governed by its physical geometry. A multimodal
system could therefore enable a new path for PIC design automation by using
the relationship between a photonic component's visual geometry and the
program that constructs it.

The crux of this relationship between the visual component and the program
that creates it is that of correct representation. Existing visual
representations, such as direct pixel traces or dense polygon arrays, can
reproduce a visual target while discarding the constructive relationships
that make its dimensions independently editable. Similarly, a prebuilt PCell
exposes parameters only within the geometric space anticipated by its author.
Correct representation hence requires an executable program that preserves
the observed geometry, records how it is constructed, and exposes
wavelength-scale dimensions as independent variables. Recent work on visual
program induction and parametric CAD has established the broader
image-to-program formulation through which visual structures are recovered as
executable, editable programs~\cite{chaudhuri,ellis2018,cadcrafter}. Beyond
this, for the photonic components specifically, the resulting program must
also encode paths, cross-sections, regions, ports, layer assignments, and
parameters that can be evaluated within a PDK.

Each candidate can also be executed, rendered at the physical scale of the
target, and compared directly with the reference geometry. Recent progress in
reinforcement learning and test-time scaling has made such verifiers integral
to model improvement as they can provide rewards during training, rank
samples, and even guide search during
inference~\cite{grpo,deepseekr1,cobbe2021verifiers,monkeys,snell}.
Fundamentally, these methods exploit an asymmetry between generation and
verification. That is, while producing a correct solution may require
substantial search, checking a completed candidate can be comparatively
inexpensive when the relevant property has a reliable test~\cite{wei}.
Rendered comparison has recently provided this signal in visual program and
parametric CAD systems~\cite{rrvf,recad}. For photonic component creation,
then, a geometric comparison provides a scalar score for selection and a
spatial residual for revision. Generation, execution, comparison, and
revision can also be repeated under a fixed compute budget. Additional
inference can be allocated to independent samples or to further revisions,
and their effects can be measured while the target, representation, and
verifier remain fixed.

\begin{figure*}[!t]
\centering
\includegraphics[width=\linewidth]{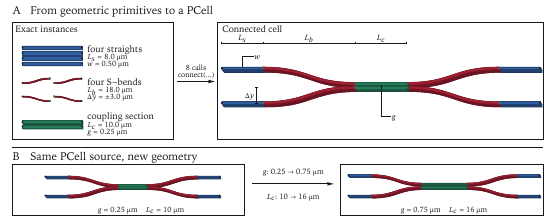}
\caption{\textbf{From vocabulary to an editable PCell.}
(a)~Four straight segments, four S-bends, and one straight coupling section
are instantiated from the catalog and aligned through eight port-connection
operations to form a complete four-port directional coupler. Hollow nodes
mark primitive endpoints before assembly and the four external ports after
assembly. Blue, red, and green identify the straight,
bend, and coupling-section provenance, independent of fabrication layers.
(b)~The same PCell source is re-executed after changing
$g$ from $0.25$ to $0.75\,\mu\mathrm{m}$ and $L_c$ from $10$ to
$16\,\mu\mathrm{m}$. The geometry changes while the four-port graph remains
fixed. All displayed geometry is generated by the pinned GDSFactory
implementation, not redrawn as an illustration.}
\label{fig:composition}
\end{figure*}

This paper makes four contributions in that direction. First, it formulates
visual-to-parametric photonic component creation through a restricted
domain-specific language (DSL) of geometric primitives, fixed-scale geometric
verification, and an explicit parametricity contract. Then, it evaluates
state-of-the-art multimodal agent configurations over eight component targets,
producing independently verified programs, while a separate iterative
campaign measures variation across samples and revision turns. Next, it
connects the recovered representation to downstream utility through
cross-stack retargeting and tiered simulation audits, showing how available
design variables and evaluator fidelity bound the resulting physical claims.
Finally, it constructs and releases an image-to-program dataset of executable
programs over the DSL, then uses that dataset and the geometric verifier to
optimize an open-weight multimodal model to produce programs in this
representation language.

\section{Setup}\label{sec:problem}

PixCell receives a binary target image with a physical footprint and a
catalog of permitted geometric operations. It works to produce an executable
program with rasterized geometry that matches its target at a fixed physical
calibration and with declared, active design parameters. The experimental
setup specifies four elements in that direction: the program representation,
target representation and calibration, a deterministic geometric verifier,
and the two model harnesses used to construct and revise candidate programs.

\begin{table*}[!t]
\centering
\caption{\textbf{The PixCell geometric language.}
Permitted constructors and operators under GDSFactory 9.20.7.
Function-relevant dimensions must remain named and editable.}
\label{tab:primitives}
\begingroup
\footnotesize
\setlength{\tabcolsep}{2.5pt}
\renewcommand{\arraystretch}{0.90}

\begin{minipage}[t]{0.49\textwidth}
\vspace{0pt}
Components and regions\par\vspace{2pt}
\begin{tabular}{@{}>{\raggedright\arraybackslash}p{0.39\linewidth}>{\raggedright\arraybackslash}p{0.56\linewidth}@{}}
\toprule
Operation & Principal live arguments \\
\midrule
\multicolumn{2}{@{}l}{\emph{Geometric regions} (\texttt{gf.components.*})} \\
\texttt{rectangle} & \texttt{size, layer, centered} \\
\texttt{circle} & \texttt{radius, layer} \\
\texttt{ring} & \texttt{radius, width, layer} \\
\texttt{cross} & \texttt{length, width, layer} \\
\texttt{triangle} & \texttt{x, y, layer} \\
\texttt{ellipse} & \texttt{radii, layer} \\
\texttt{regular\_polygon} & \texttt{sides, side\_length, layer} \\
\texttt{L} & \texttt{width, size, layer} \\
\midrule
\multicolumn{2}{@{}l}{\emph{Port-bearing components} (\texttt{gf.components.*})} \\
\texttt{straight} & \texttt{length, cross\_section} \\
\texttt{bend\_euler} & \texttt{radius, angle, p, cross\_section} \\
\texttt{bend\_circular} & \texttt{radius, angle, cross\_section} \\
\texttt{bend\_s} & \texttt{size, cross\_section} \\
\texttt{bezier} & \texttt{control\_points, cross\_section} \\
\texttt{taper} & \texttt{length, width1, width2} \\
\texttt{coupler\_straight} & \texttt{length, gap, cross\_section} \\
\midrule
\multicolumn{2}{@{}l}{\emph{Component container and instances}} \\
\texttt{gf.Component} & \texttt{()} \\
\texttt{c.add\_ref}$^{\dagger}$ & \texttt{component, columns, rows, column\_pitch, row\_pitch} \\
\texttt{c.add\_port} & \texttt{name, port} \\
\bottomrule
\end{tabular}
\end{minipage}
\hfill
\begin{minipage}[t]{0.49\textwidth}
\vspace{0pt}
Paths and operators\par\vspace{2pt}
\begin{tabular}{@{}>{\raggedright\arraybackslash}p{0.40\linewidth}>{\raggedright\arraybackslash}p{0.55\linewidth}@{}}
\toprule
Operation & Principal live arguments \\
\midrule
\multicolumn{2}{@{}l}{\emph{Cross-section and paths}} \\
\multicolumn{2}{@{}l}{\textit{API:} \texttt{gf.cross\_section.cross\_section}; \texttt{gf.Path}; \texttt{gf.path.*}} \\
\texttt{cross\_section} & \texttt{width, layer} \\
\texttt{Path} & \texttt{points} \\
\texttt{straight} & \texttt{length} \\
\texttt{arc} & \texttt{radius, angle} \\
\texttt{euler} & \texttt{radius, angle, p} \\
\texttt{smooth} & \texttt{points, radius} \\
\texttt{spiral\_archimedean} & \texttt{min\_bend\_radius, separation, number\_of\_loops, npoints} \\
\texttt{path.append} & \texttt{segment} \\
\midrule
\multicolumn{2}{@{}l}{\emph{Body and set operations} (\texttt{gf.path.*}; \texttt{gf.boolean})} \\
\texttt{extrude} & \texttt{p=path, cross\_section=xs} \\
\texttt{transition} & \texttt{cross\_section1, cross\_section2, width\_type} \\
\texttt{extrude\_transition} & \texttt{p=path, transition} \\
\texttt{boolean} & \texttt{A, B, operation, layer} \\
\midrule
\multicolumn{2}{@{}l}{\emph{Reference operations} (\texttt{ref.*})} \\
\texttt{connect} & \texttt{port, destination} \\
\texttt{move} & \texttt{origin, destination} \\
\texttt{rotate} & \texttt{angle, center} \\
\texttt{mirror} & \texttt{p1, p2} \\
\midrule
\multicolumn{2}{@{}l}{\emph{Routing} (\texttt{gf.routing.*})} \\
\texttt{route\_single}$^{\ddagger}$ & \texttt{component, port1, port2, cross\_section} \\
\bottomrule
\end{tabular}
\end{minipage}

\vspace{2.5pt}
\begin{minipage}[t]{0.49\textwidth}
\raggedright\footnotesize
$^{\dagger}$\,\texttt{add\_ref} creates a single reference by default and an
array when \texttt{columns} or \texttt{rows} exceed one.
\end{minipage}
\hfill
\begin{minipage}[t]{0.49\textwidth}
\raggedright\footnotesize
$^{\ddagger}$\,\texttt{route\_single} adds route geometry directly to its
parent \texttt{Component} and returns a \texttt{ManhattanRoute}.
\end{minipage}
\endgroup

\end{table*}

\subsection{Geometric primitive representation}\label{sec:representation}

The outputs here are Python programs that construct a GDSFactory component
from the vocabulary in Fig.~\ref{fig:primitives} and
Table~\ref{tab:primitives}~\cite{gdsfactory}. The language contains geometric
regions, port-bearing waveguide elements, parameterized paths and
cross-sections, Boolean operations, and reference-based composition.
Moreover, the DSL's path system keeps the centerline and cross-section
independently editable, while ports define how instantiated elements connect.
Dimensions such as widths, gaps, radii, taper lengths, and coupling lengths
also appear as named top-level variables. As a result, we have a compact
geometric DSL~\cite{chaudhuri} whose programs can be executed and audited.
The source contract also rejects raw polygon or vertex emission and pre-built
device cells outside the permitted DSL vocabulary. A pixel trace may overlap
the target without recovering constructive parameters, while a pre-built
cell substitutes retrieval for component authoring. The programs instead
must use and instantiate primitives, extrude paths, apply Boolean operations,
position references, and connect compatible ports.
Figure~\ref{fig:composition} illustrates these operations and the resulting
program structure, in which changing parameters changes the executed
geometry.

\begin{figure*}[t]
\centering
\includegraphics[width=\linewidth]{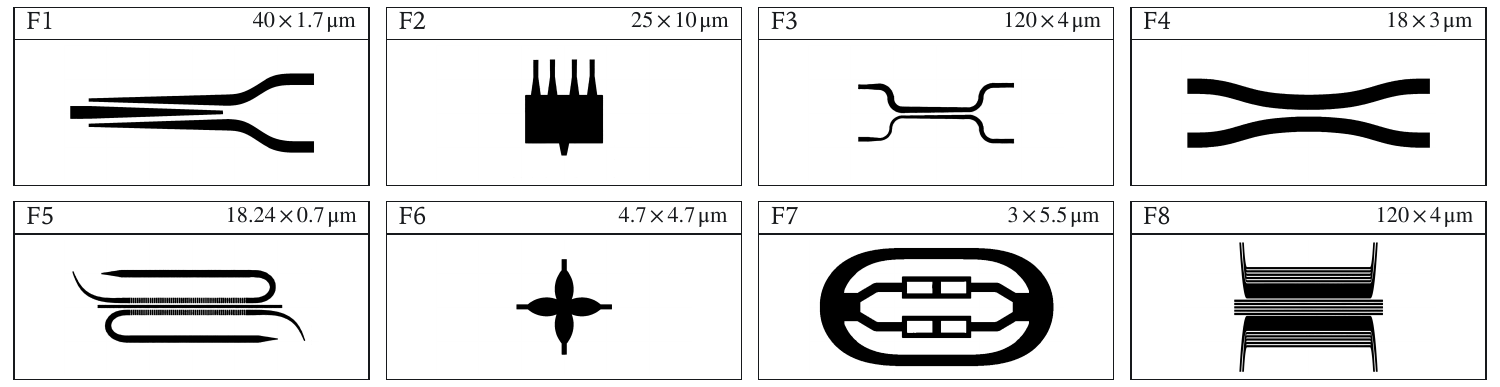}
\caption{\textbf{Benchmark targets.} Each device, F1--F8, is presented as a
binary silhouette and physical footprint; panels are not shown at a common
scale. Source attributions are: F1, 3-dB adiabatic
splitter~\cite{splitter2020}; F2, $1\times4$ MMI
splitter~\cite{malka2016}; F3, adiabatic $2\times2$
coupler~\cite{mao2019}; F4, thin-film lithium niobate directional
coupler~\cite{huang2023}; F5, device from a microring-like resonator
system~\cite{nanobeam}; F6, ultra-compact waveguide
crossing~\cite{crossing2020}; F7, ring-based WDM filter
section~\cite{wdmfilter}; F8 is retained as a provenance-incomplete target
with its extraction record included in the release metadata.}
\label{fig:targets}
\end{figure*}

\subsection{Targets and calibration}

\begin{figure}[!t]
\centering
\includegraphics[width=\linewidth]{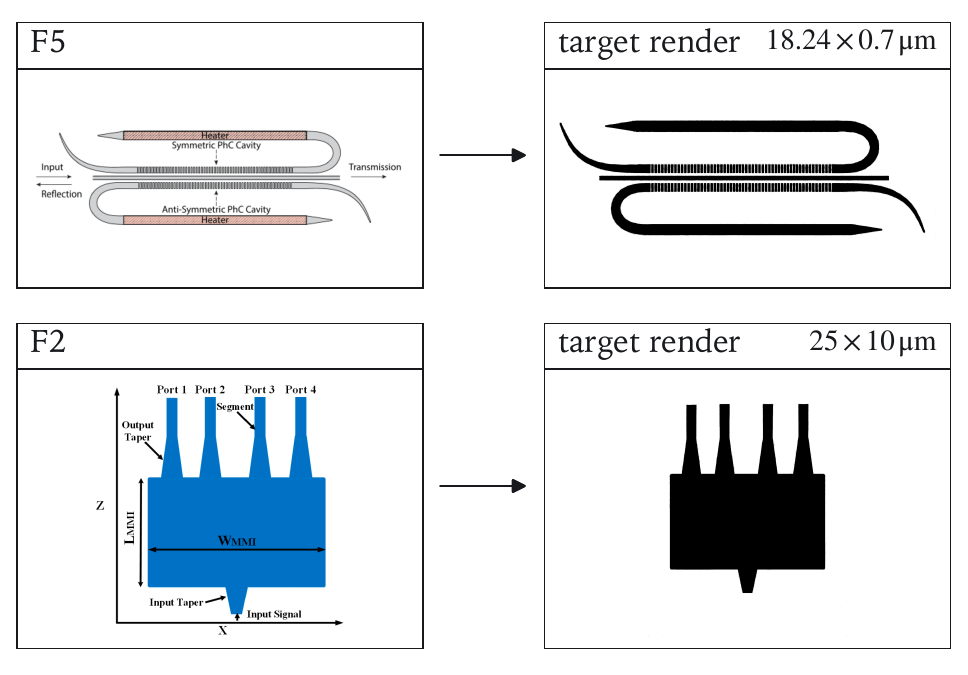}
\caption{\textbf{Preparation of target devices.} Source figure
panels are standardized once into frozen masks and physical footprints.
F5 shows panel isolation from a compound figure; F2 shows annotation removal
while preserving topology. The resulting inputs are fixed before model
evaluation.}
\label{fig:extraction}
\end{figure}

The benchmark contains eight photonic components, F1--F8
(Fig.~\ref{fig:targets}), which span smooth adiabatic paths, compact
multiport regions, resonant loops, crossings, and repeated fine features.
The physical footprints of these devices range from a $4.7\,\mu$m square to
a $120\,\mu$m-long bus, and each model receives the raster and footprint with
device identity, source text, captions, and the other benchmark cells
remaining hidden. However, the published panels often include annotations
or surrounding page content, and so the selected panel is isolated and
converted once into a device-only binary mask, with dark pixels denoting
material on the waveguiding layer. Gemini~3 Pro Image produces this canonical
rendering~\cite{gemini3image}, and a separate Gemini~3 Pro vision pass
extracts the physical footprint from the source text or scale
annotations~\cite{gemini3}. The image masks and footprint calibration records
are then frozen and reused across every configuration
(Fig.~\ref{fig:extraction}). All reported results use GDSFactory 9.20.7 and a
Python environment with runtime dependencies and process instructions
included in the released repository.

\subsection{Verification}\label{sec:verifier}

The geometric verifier executes each candidate program by rendering its
output at target-derived physical calibration and comparing the result with
the reference target. For each of these targets, the foreground bounding box
and stated footprint define separate horizontal and vertical
pixel-per-micron scales
$\boldsymbol{\kappa}=(\kappa_x,\kappa_y)$. The rendering preserves these
scales, aligns the candidate's left edge with the target, and also centers it
vertically using its own bounds. Importantly, the candidate geometry is never
rescaled to fit the target.

A candidate is represented as $(p,\theta)$, where $p$ is the program
structure and $\theta\in\mathbb{R}^{k(p)}$ contains its named dimensions.
The rendered candidate is
\begin{equation}\label{eq:render}
\hat y(p,\theta)\;=\;R_{\boldsymbol{\kappa}}\!\bigl(\Exec(p,\theta)\bigr)
\;\in\;\{0,1\}^{H\times W},
\end{equation}
and its geometric agreement with target $y$ is
\begin{equation}\label{eq:iou}
\iou(y,\hat y)\;=\;\frac{|\,y\wedge\hat y\,|}{|\,y\vee\hat y\,|},
\end{equation}
where intersection and union are evaluated over the foreground pixels.
Importantly, IoU is the geometric acceptance metric used throughout the
reconstruction processes, but the verifier also reports Dice overlap and
binary sum of squared errors (SSE).

\begin{figure*}[t]
\centering
\includegraphics[width=\linewidth]{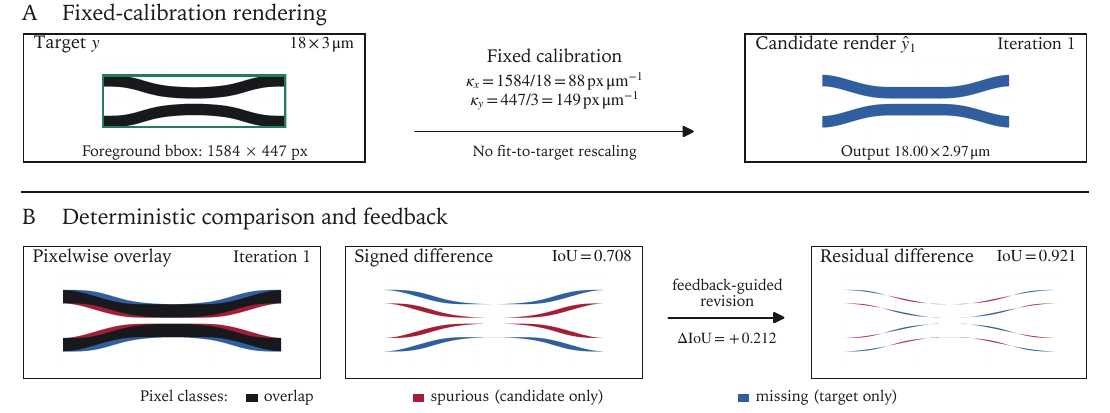}
\caption{\textbf{Geometric verification for one benchmark cell.} The
F4 target has footprint $18\times3\,\mu$m and fixed axis calibrations of
$88$ and $149\,\mathrm{px}/\mu$m. The candidate is rendered without
fit-to-target rescaling. Black marks overlap, red candidate-only material,
and blue target-only material. One feedback-guided revision raises IoU from
$0.708$ to $0.921$.}
\label{fig:verification}
\end{figure*}

For a configuration with $n$ cells, source-compliant geometric performance
is reported as usable IoU,
\begin{equation}\label{eq:usable}
U=\frac{1}{n}\sum_{i=1}^{n}s_i\,\iou_i
=\Pr(s=1)\,\mathbb{E}\!\left[\iou\mid s=1\right],
\end{equation}
where $s_i=1$ when the program passes the source contract and zero otherwise.
Thus every source violation contributes zero without hiding its raw geometric
score.

\subsection{Acceptance gates and evaluation harnesses}
\label{sec:parametricity}

Following this, scores are recomputed and recorded from the archived program,
GDS, target, and calibration. Analysis scripts reject prohibited device-level
calls and raw polygon emission; the resulting verdict also determines $s_i$ in
Eq.~\eqref{eq:usable}. Parametric ability is measured separately by
perturbing a top-level variable, rebuilding the program, and confirming that
the executed geometry changes. This establishes that selected variables
affect geometry without testing their independence. Parametricity is recorded
separately and does not alter $U$. These experiments use two harnesses. The
first is an iterative API harness in the form of an external pipeline that
controls sampling, execution, selection, and feedback. In the coding-agent
harness, on the other hand, the model controls its own execute-measure-revise
loop. Both use the same frozen F1--F8 targets, representation contract, and
geometric verifier, while their prompts, tool access, and aggregation differ.

\begin{figure*}[t]
\centering
\includegraphics[width=\linewidth]{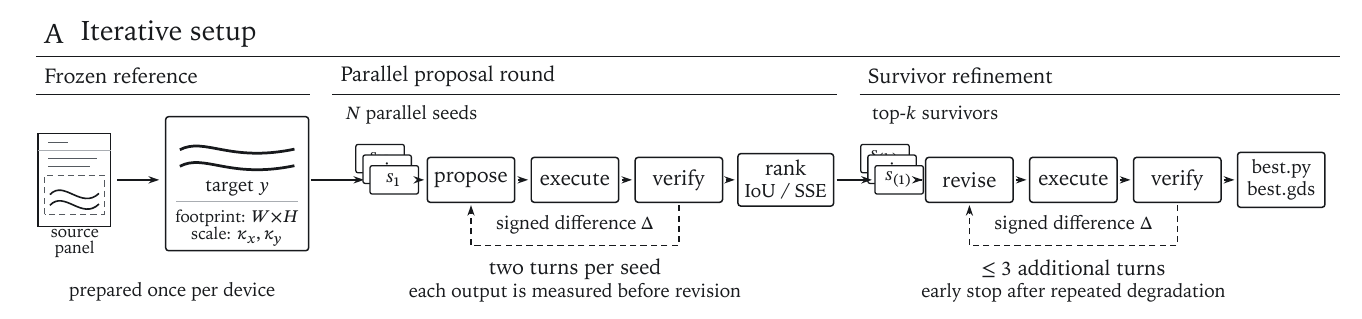}
\par\vspace{0.4em}
\includegraphics[width=\linewidth]{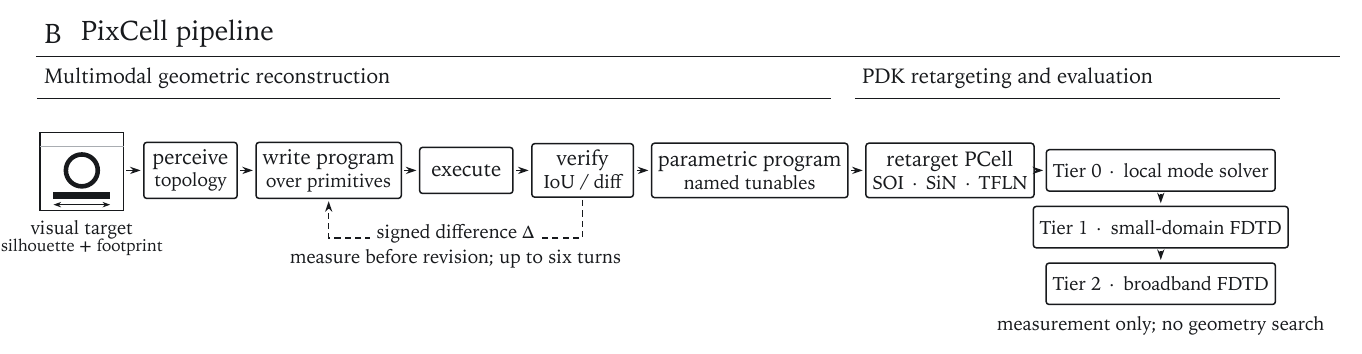}
\caption{\textbf{Evaluation harnesses and PixCell pipeline.}
(a)~The iterative API harness samples programs in parallel, ranks them with
the verifier, and continues selected candidates with spatial feedback.
(b)~A coding agent constructs and revises a parametric program using the
same geometric verifier. Accepted parameters later connect the program to
tiered physical evaluation.}
\label{fig:phases}
\end{figure*}

\paragraph{Iterative API harness.}\label{sec:iterative}

In the iterative API process, the model acts as a stochastic program
generator while the pipeline controls the loop
(Fig.~\ref{fig:phases}(a)). Its prompt contains the target raster image,
footprint, and primitive catalog without any device name or worked example.
Each device then has 5 to 25 independent samples, each with two initial
proposal-execution-verification turns. The verifier ranks candidates, after
which selected seeds may receive up to three further difference-guided turns.

\paragraph{Coding-agent harness.}\label{sec:agentsetup}

In the coding-agent harness, each benchmark cell is assigned to one CLI
coding task~\cite{claudecode,codexcli}. The matrix contains 26 configurations
and 208 programs covering all eight targets. Each target-specific task
directory begins with the silhouette and footprint, and every worker can use
the repository's Python environment and geometric verifier to write and
execute its own program there. At run time, the agent can also inspect its
own raster and compare images to the target, and decide how to revise the
candidate. These runs are limited to six rounds, and one final program is
retained without multi-seed selection. The worker briefs prohibit access to
other cells, prior campaign outputs, and benchmark metadata and are identical
across cells except for nomenclature requirements. The conductor confirms
each requested model and reasoning setting by status-line readback, monitors
machine-readable completion markers, and archives programs, transcripts,
intermediate renders, and final measurements. In this way, only the
conductor-verified artifact measurements enter the record.

\section{Results}\label{sec:results}

Our two harnesses measure complementary uses of the same verifier. While the
iterative API campaign measures selection across seeds and revision turns,
the coding-agent campaign measures one retained program in every cell of a
balanced configuration-by-target matrix. Raw IoU reports geometric agreement
and $U$ additionally assigns zero to source-violating programs.

\subsection{Verifier-guided sampling and revision}\label{sec:june}

The iterative API record contains 63 experiments with multiple seeds. The
mean difference between the best and worst seed is $0.246$ IoU, and $84\%$
of experiments span at least $0.10$. The first seed is best in only
$12.7\%$ of experiments. Across 640 revision trajectories, mean first-turn
IoU is $0.369$, while the best observed turn averages $0.416$; $65.9\%$ of
trajectories peak after the first turn. Relative to retaining turn one,
best-of-two, best-of-three, and best-of-five selection improve mean IoU by
$10.0\%$, $16.0\%$, and $18.3\%$, respectively.

\begin{figure*}[t]
\centering
\includegraphics[width=\linewidth]{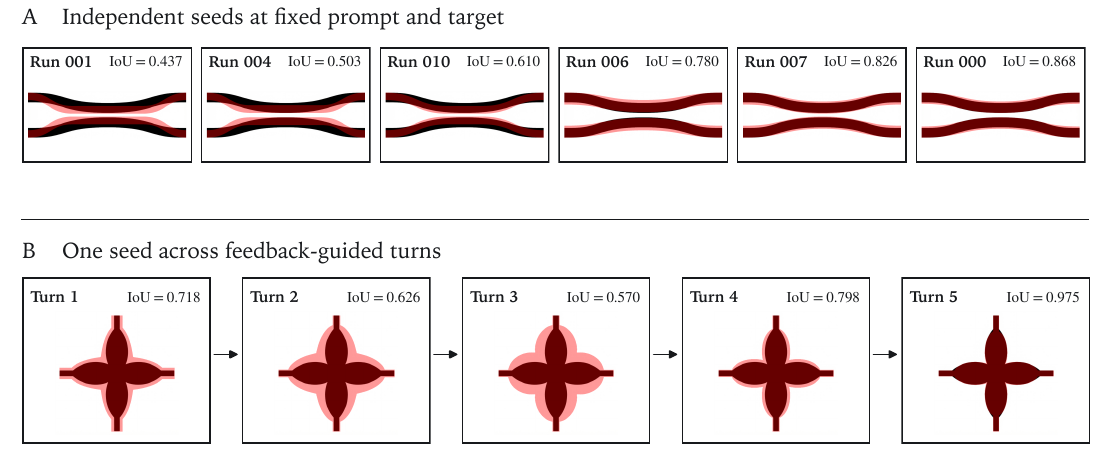}
\caption{\textbf{Sampling and revision in the iterative API harness.}
(a)~Six archived Gemini-3-Pro seeds generated from the same prompt and F4
target span $\iou=0.44$--$0.87$. Maroon denotes overlap, pink
candidate-only material, and black target-only material.
(b)~One F6 seed over five verification-guided turns follows
$0.72\to0.63\to0.57\to0.80\to0.975$.}
\label{fig:juneseeds}
\end{figure*}

The same pattern is visible in individual artifacts
(Fig.~\ref{fig:juneseeds}). Eleven Gemini-3-Pro seeds on F4 range from
$0.360$ to $0.868$. One F6 seed follows the non-monotone sequence
$0.72\to0.63\to0.57\to0.80\to0.975$. Verification therefore supplies both
the ranking needed to allocate inference and the spatial residual needed to
revise a candidate. Despite these gains from the selection and revision
process, the mean best-turn IoU remained $0.416$, and the external pipeline
was compelled to coordinate multiple samples, execution, ranking, and
feedback. While this direction could have been furthered, we generally found
that the large number of seeds and iterations needed to make meaningful
progress across these targets still produced only mediocre scores and, more
importantly, was inefficient in both tokens and time. This motivated the
coding-agent campaign, in which each model controls its own
execute-measure-revise loop through the same verifier, with the data in the
benchmark (Fig.~\ref{fig:emergence}) validating this direction.

\subsection{Blind coding-agent benchmark}\label{sec:agents}

\begin{figure*}[t]
\centering
\includegraphics[width=\linewidth]{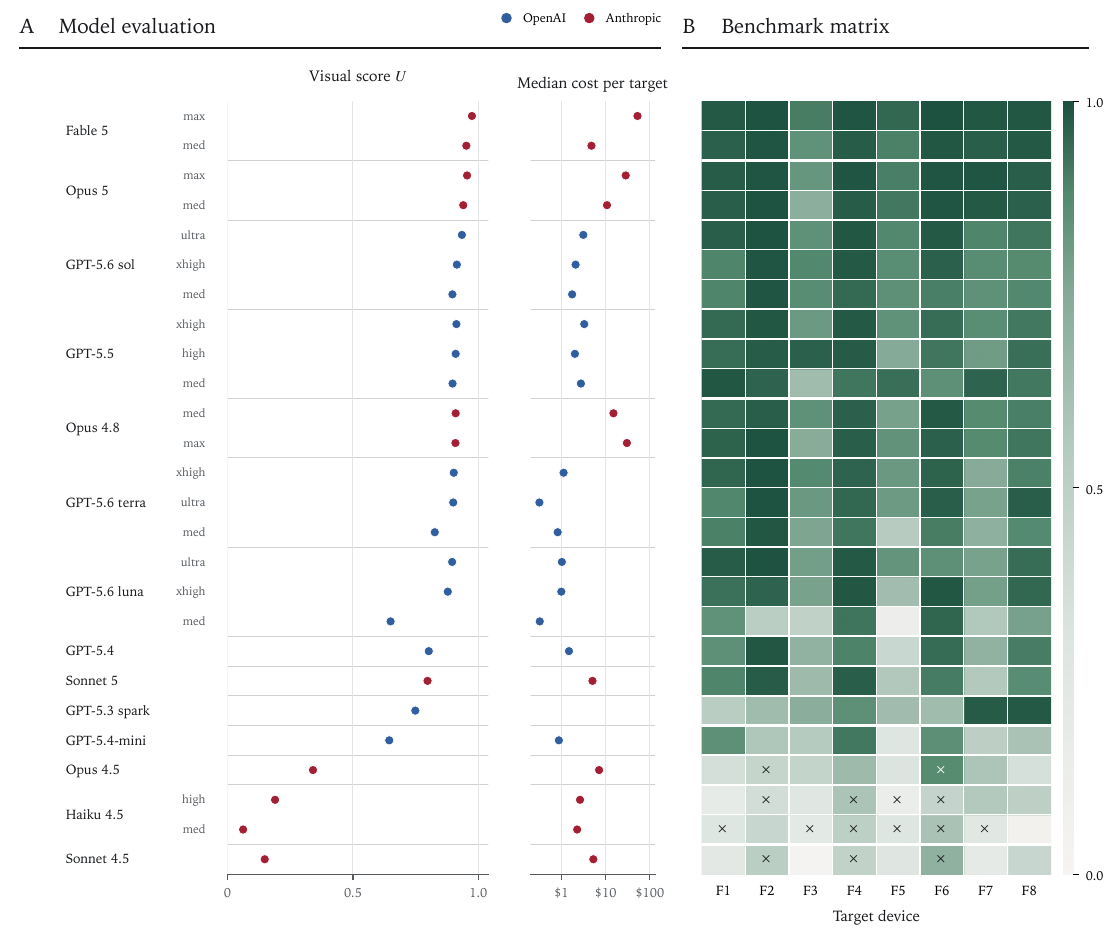}
\caption{\textbf{Coding-agent benchmark.}
(a)~Usable IoU $U$ and median API-list-price-equivalent cost per target for
26 configurations and 208 programs. Cost converts measured tokens at the
published rates frozen on July 27, 2026; one unpriced preview configuration
is omitted from the cost axis. Blue denotes OpenAI and red Anthropic.
(b)~Independently recomputed raw IoU across F1--F8 in the same row order.
Each $\times$ marks a source violation whose visible geometric score
contributes zero to $U$.}
\label{fig:emergence}
\end{figure*}

The balanced benchmark evaluates 26 model and reasoning configurations on
all eight targets, producing 208 independently rescored programs
(Fig.~\ref{fig:emergence}). Fable~5 at max effort reaches the highest mean
IoU, $0.974$, followed by Opus~5 at max effort with $0.955$ and Fable~5 at
medium effort with $0.952$. All three are source-compliant in every cell, so
their raw means equal $U$. Across the full matrix, twenty-two of the 26
configurations are source-compliant on all eight targets, so each has raw
mean IoU equal to $U$.

\begin{figure*}[t]
\centering
\includegraphics[width=\linewidth]{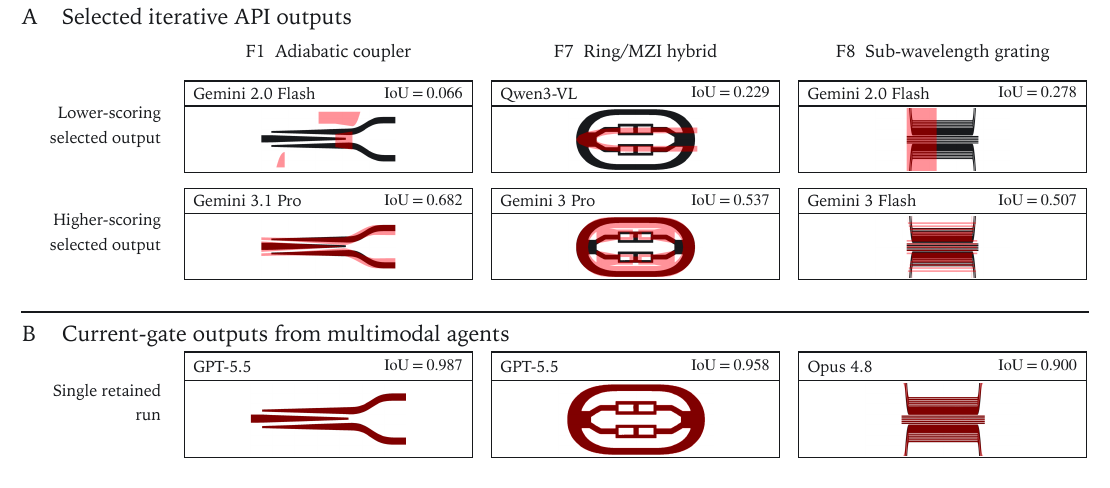}
\caption{\textbf{Selected reconstructions.}
(a)~Lower- and higher-scoring outputs from the iterative API record.
(b)~One gate-passing coding-agent run for each displayed target, with
independently recomputed IoU. Maroon denotes overlap, pink candidate-only
material, and black target-only material.}
\label{fig:beforeafter}
\end{figure*}

\subsection{Configuration and target effects}\label{sec:capability}

Each configuration was evaluated once on each of eight targets, so the
benchmark does not estimate run-to-run variability. The reported bootstrap
intervals resample targets and therefore summarize sensitivity to target
choice without claiming any stochastic uncertainty within a
configuration-target pair. Rather than focusing on the depth of the benchmark
and the extent of figures covered, we felt that the more important question
was whether we could prove that such capabilities had emerged in these
multimodal agents. In particular, we want to focus on what these results imply
for systems that can visually create parametric components, both in terms of
PDK retargeting and the future of training models for photonic design
automation. For the results in Fig.~\ref{fig:emergence}, the $95\%$ intervals
are $0.974~[0.952,0.991]$ for Fable~5 max and
$0.955~[0.914,0.987]$ for Opus~5 max.

Performance varies substantially by target (Fig.~\ref{fig:emergence}(b)). F3
and F5 remain the lowest-mean targets at $0.695$ and $0.660$. F3 spans
$0.019$--$0.967$ across configurations, while F8 spans $0.058$--$0.986$.
No configuration wins all eight targets: the highest-mean configuration
wins four, while GPT-5.5 medium, GPT-5.5 high, and Opus~5 max lead at least
one of the others.

Higher configured reasoning effort is associated with higher mean IoU in
most within-family ladders, although the relation is not uniformly
monotone. Fable~5 rises from $0.952$ at medium effort to $0.974$ at max,
and GPT-5.6 sol rises from $0.896$ at medium to $0.934$ at ultra. Opus~4.8
is effectively unchanged, moving from $0.910$ at medium to $0.909$ at max,
while GPT-5.6 terra peaks at xhigh and declines slightly at ultra.

Source compliance separates several configurations with similar raw
geometric scores. In total, 193 of 208 programs satisfy the source gate.
The 15 violations occur in four configurations: 2 of 8 Opus-4.5 cells, 3 of
8 Sonnet-4.5 cells, 4 of 8 Haiku-4.5-high cells, and 6 of 8 Haiku-4.5 medium
cells. Every flagged program contains prohibited \texttt{add\_polygon} calls.
Fundamentally, however, the coding-agent results are not Pass@$k$
benchmarking results. We again wish to underscore that the results in this
section focus less on superiority within model classes than on the emergence
of this intelligence capability and the questions motivated by such an
ability. As a result, Section~\ref{sec:pdk} focuses on whether these retained
variables support process-specific physical retargeting as a viable aspect of
such multimodal photonic design automation.

\subsection{Runtime and cost}\label{sec:inference}

The cost axis in Fig.~\ref{fig:emergence}(a) converts measured token buckets
to the published list-price equivalent under one frozen rate
card~\cite{openai_pricing,anthropic_pricing}. The tested frontier spans more
than two orders of magnitude, with GPT-5.6 terra ultra reaching $U=0.900$ at
a median $\$0.317$ per target, GPT-5.6 sol ultra reaching $0.934$ at
$\$3.121$, Fable~5 medium reaching $0.952$ at $\$4.766$, Fable~5 at max
reaching $0.974$ at $\$52.452$, and Opus~5 max reaching $0.955$ at
$\$28.405$. Additional compute is therefore one route to higher performance,
but model choice and target geometry materially affect the return.

\section{Retargeting across stack models}\label{sec:pdk}

While geometric verification establishes an executable representation and
the benchmarks above have shown how state of the art multimodal intelligence
is able to understand the representation and recreate a component with
accurate parameters, physical evaluation is still limited. For this, we set
up a tiered simulation framework. Tier~0 uses local mode solutions to compute
effective and group indices, path-delay quantities, and coupled-mode
estimates. Tier~1 uses capped small-domain FDTD for transmission. Tier~2 uses
human-authorized broadband FDTD for full-device audits~\cite{tidy3d}. The
evaluators here measure functional quantities while the agent decides how to
modify the program. Experiments here use explicit models of 220-nm SOI,
400-nm SiN, and 400-nm TFLN. A pass requires the implemented optical figure
of merit and footprint constraint to hold on the selected stack. The TFLN
model is isotropic, and the gate does not enforce the bend-radius and
guide-spacing fields or a complete foundry rule deck. The results therefore
measure functional reachability under these stack models.

\subsection{Editable routing enables cross-stack retargeting}
\label{sec:ladder}

Retargetability is the set of valid geometries reachable through a
program's live parameter variables. A library PCell exposes the coordinates
chosen by its author. A reconstructed program exposes those retained by the
coding agent. The parameter ranges, evaluator, and geometric constraints
determine which operating points either representation can reach.

The first test starts from an unbalanced Mach--Zehnder interferometer (MZI)
reconstructed independently by Opus-4.8 medium and GPT-5.5 xhigh. Each
worker receives the calibrated silhouette and its
$91.1\times50.75\,\mu\mathrm{m}^{2}$ footprint. Retargeting requests a
free spectral range (FSR) of $8.0$\,nm within $\pm2\%$ at $1550$\,nm on
each stack. The tier-0 evaluator combines a local group-index solution with
path lengths from the executed program.

The as-pictured MZI has a $10.0\,\mu$m imbalance, giving modeled FSRs of
$59.21$\,nm on SOI, $131.37$\,nm on SiN, and $114.48$\,nm on TFLN.
The native imbalance variable of the GDSFactory MZI reaches the spectral
target, but realizes the extra path length through vertical extension. Its
minimum tested heights are $58.8$, $96.5$, and $89.9\,\mu$m, all above the
$50.75\,\mu$m budget.

The reconstructed programs can revise their primitive routing. Both agents
replace the long-arm geometry with a folded route and use imbalances of
approximately $74.0$, $148.6$, and $134.8\,\mu$m across SOI, SiN, and
TFLN. All six agent-by-stack designs satisfy the analytic FSR and footprint
gates, with FSRs from $7.9995$ to $8.0007$\,nm. This test compares editable
primitive code with the native interface of a library PCell. The
fixed-interface experiment in Sec.~\ref{sec:catalog} removes that authoring
asymmetry.

\begin{figure}[!t]
\centering
\includegraphics[width=\linewidth]{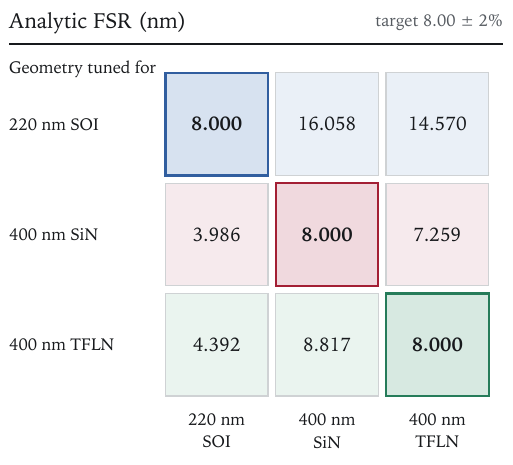}
\caption{\textbf{Cross-stack transfer of the retuned MZI.} Each row fixes
one Opus-4.8-medium geometry and evaluates it unchanged on every column
stack. Outlined diagonal cells meet the $8.0$\,nm $\pm2\%$ target; all six
off-diagonal transfers miss. The GPT-5.5-xhigh matrix agrees within
$0.002$\,nm per cell.}
\label{fig:crossmatrix}
\end{figure}

A geometry tuned on one stack does not transfer unchanged to the other two.
Figure~\ref{fig:crossmatrix} evaluates each Opus-tuned geometry on every
stack model. The diagonal entries pass, whereas all six off-diagonal
entries miss the $\pm2\%$ interval, by $9.3\%$ to $100.7\%$. The
corresponding GPT-5.5 matrix differs by less than $0.002$\,nm in every
cell. The same pattern holds for both agents: all 12 off-diagonal polygon
transfers fail, while the executable program can be retargeted.

\subsection{Fixed-representation comparison}\label{sec:catalog}

A second experiment fixes both representations for five catalog devices:
an MZI, ring resonator, directional coupler, delay spiral, and distributed
Bragg reflector. GPT-5.5 high and Opus-4.8 medium first reconstruct each
catalog image using primitives. Retargeting then operates either on the
reconstructed program or on a wrapper around the corresponding GDSFactory
PCell. Both arms may change existing parameters but may not rewrite their
code. The library PCell's modeled home-SOI behavior defines the target
requested on all three stacks, giving 60 total outcomes.

\begin{table}[!b]
\caption{\textbf{Fixed-representation retargeting.} R: reconstructed
primitive program; L: library PCell. Each target is the library PCell's
modeled home-SOI behavior, requested on 220-nm SOI, 400-nm SiN, and 400-nm
TFLN. Filled and open markers denote pass and observed miss under the
implemented tier-0 figure-of-merit and footprint gate.}
\label{tab:twoarm}
\begingroup
\newcommand{\retargetpass}{\textcolor[HTML]{246B5A}{\raisebox{0.08ex}{\scalebox{1.05}{$\bullet$}}}}
\newcommand{\retargetmiss}{\textcolor[HTML]{7E858A}{\raisebox{0.08ex}{\scalebox{1.05}{$\circ$}}}}
\footnotesize
\setlength{\tabcolsep}{3.35pt}
\renewcommand{\arraystretch}{1.06}
\begin{ruledtabular}
\begin{tabular}{@{}llccc@{\hspace{3.5pt}}ccc@{}}
& & \multicolumn{3}{c}{GPT-5.5 high} & \multicolumn{3}{c}{Opus-4.8 medium} \\
Device & Rep. & SOI & SiN & TFLN & SOI & SiN & TFLN \\
\colrule
MZI     & R & \retargetpass & \retargetpass & \retargetpass & \retargetpass & \retargetpass & \retargetpass \\
        & L & \retargetpass & \retargetmiss & \retargetpass & \retargetpass & \retargetmiss & \retargetpass \\
\addlinespace[0.12em]
Coupler & R & \retargetpass & \retargetpass & \retargetpass & \retargetpass & \retargetpass & \retargetmiss \\
        & L & \retargetpass & \retargetmiss & \retargetpass & \retargetpass & \retargetpass & \retargetmiss \\
\addlinespace[0.12em]
DBR     & R & \retargetpass & \retargetpass & \retargetpass & \retargetpass & \retargetmiss & \retargetmiss \\
        & L & \retargetpass & \retargetpass & \retargetpass & \retargetpass & \retargetpass & \retargetpass \\
\addlinespace[0.12em]
Spiral  & R & \retargetpass & \retargetmiss & \retargetmiss & \retargetpass & \retargetpass & \retargetpass \\
        & L & \retargetpass & \retargetpass & \retargetpass & \retargetpass & \retargetpass & \retargetpass \\
\addlinespace[0.12em]
Ring    & R & \retargetpass & \retargetmiss & \retargetmiss & \retargetpass & \retargetmiss & \retargetmiss \\
        & L & \retargetpass & \retargetmiss & \retargetmiss & \retargetpass & \retargetmiss & \retargetmiss \\
\colrule
Total passes & R & \multicolumn{3}{c}{11/15} & \multicolumn{3}{c}{10/15} \\
             & L & \multicolumn{3}{c}{11/15} & \multicolumn{3}{c}{11/15} \\
\end{tabular}
\end{ruledtabular}
\endgroup
\end{table}

Across the 30 stack-by-model cases for each representation, reconstructed
programs pass 21 and library PCells pass 22. The paired verdict agrees in
23 cases. Among the seven disagreements, the reconstructed program passes
three times when the PCell misses, and the PCell passes four times when the
reconstruction misses. Neither representation class dominates.

The individual disagreements show why the aggregate is balanced. Both
reconstructed MZIs retain sufficient routing freedom to satisfy the SiN
target within the footprint. The library MZI reaches the FSR target but is
too large. Conversely, one reconstructed spiral fixes its loop
count and misses the SiN and TFLN delay targets; the other reconstruction
keeps that coordinate live and passes. For the ring, both representations
miss on SiN and TFLN because the radius required by the modeled FSR exceeds
the footprint. The 21--22 near tie shows no aggregate representation-class
advantage in this experiment. Success tracks the live coordinates retained
by each program.

\begin{figure*}[t]
\centering
\includegraphics[width=\linewidth]{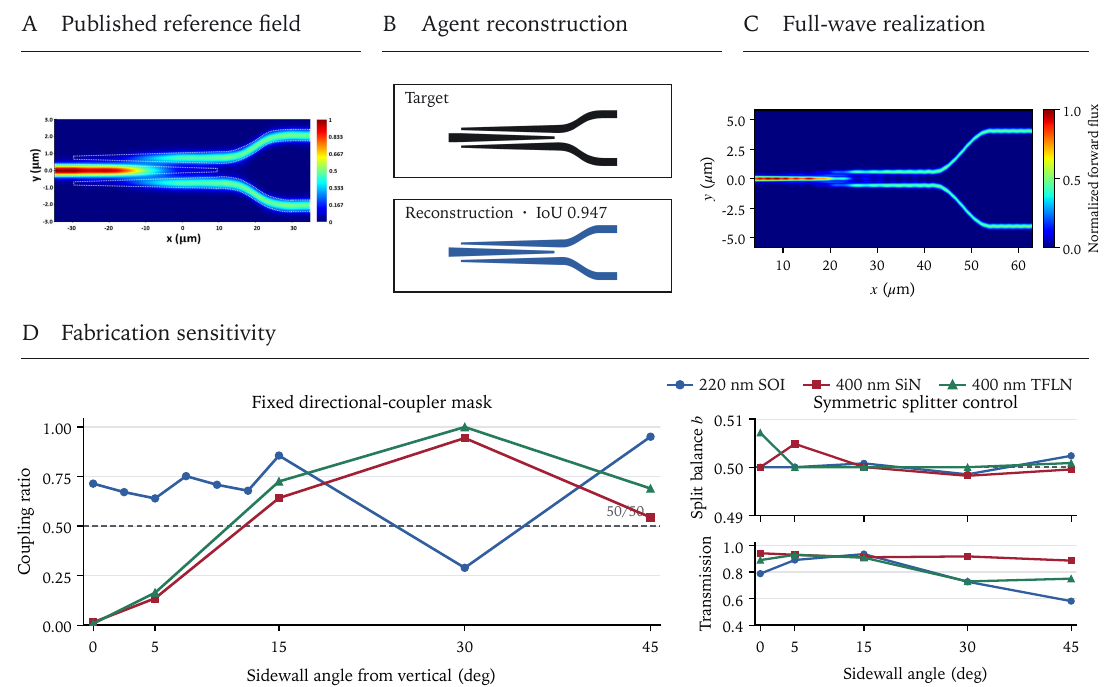}
\caption{\textbf{Full-wave function and fabrication sensitivity.}
(a)~Published field intensity for the shallow-etched rib splitter of Nguyen
\emph{et al.}~\cite{splitter2020}, reproduced under CC BY 4.0.
(b)~The silhouette supplied to the worker and its executed primitive
reconstruction.
(c)~Forward Poynting-flux density for the SOI-retargeted realization. Mode
monitors give transmission $T=0.7873$ and split balance $b=0.500001$.
Panels (a) and (c) differ in geometry, observable, domain, material model,
and normalization, so their comparison is qualitative.
(d)~Left: coupling ratio of a fixed reconstructed directional-coupler mask
under uniform sidewall tilt. Right: split balance and transmission for
stack-specific, mirror-symmetric F1 splitter controls. Markers are simulated
samples and connecting segments guide the eye.}
\label{fig:sidewall}
\end{figure*}

\subsection{Full-wave and fabrication audits}
\label{sec:third}\label{sec:release}\label{sec:channel}

Full-wave simulation first audits the inexpensive analytic tier. A broadband
FDTD run of the Opus-retargeted SOI MZI gives a fringe spacing of
$7.77$\,nm, compared with $7.9995$\,nm from the analytic evaluator, a
$2.9\%$ difference. The simulated fringe contrast is $12.24$\,dB.
Separate tier-1 splitter fixtures measure $2.012$\,dB per Opus
reconstruction, $2.994$\,dB per GPT-5.5 reconstruction, and
$1.075$\,dB per generic PCell. The corresponding two-splitter estimates,
$4.024$, $5.989$, and $2.15$\,dB, all exceed the specified $1$\,dB
ceiling; these estimates are not full-device broadband loss measurements.

The same analytic tier fails quantitatively for the directional coupler.
For the reconstructed SiN coupler, the parallel-guide supermode model
predicts a coupling ratio of $0.9439$, while a matched full-wave fixture
gives $0.1711$, a $5.5\times$ overestimate. The full-wave geometry includes
the access and S-bend regions omitted from the analytic model. The analytic
evaluator differs from the full-wave result by $2.9\%$ for the tested MZI
fringe spacing and overestimates this coupler ratio by $5.5\times$. Coupler
entries in Table~\ref{tab:twoarm} are tier-0 outcomes.

A separate MMI fixture shows the value of stack-specific redimensioning.
The generic PCell measures $1.08$\,dB per splitter on SOI. Holding its
$2.5\times5.5\,\mu\mathrm{m}^{2}$ body fixed while adapting only the access
width to SiN raises loss to $3.06$\,dB. A primitive MMI dimensioned for the
SiN mode solution measures $1.09$\,dB per splitter.

F1 provides an end-to-end check. The worker received only the standardized
silhouette and its $40.0\times1.7\,\mu\mathrm{m}^{2}$ footprint, not the source
text or published field panel. Its primitives-only program achieved
an IoU of $0.947$. Retargeting then derived a simplified 220-nm SOI realization
from that program. At $1550$\,nm, full-wave simulation yields symmetric
two-output flow, split balance $0.500001$, and monitored transmission
$0.7873$ (Fig.~\ref{fig:sidewall}(a)--(c)). This establishes the intended
splitting function for the simplified SOI realization; the published
shallow-etched rib field provides a qualitative reference~\cite{splitter2020}.

A two-dimensional mask also omits fabrication variables. We hold one
Opus-reconstructed directional-coupler mask fixed and vary uniform sidewall
tilt in full-wave simulations. On SiN, the coupling ratio changes from
$0.014$ at vertical walls to $0.944$ at $30^{\circ}$; on TFLN it changes
from $0.008$ to $1.000$. The first interpolated $50/50$ crossings occur at
$12.2^{\circ}$ and $11.0^{\circ}$, respectively
(Fig.~\ref{fig:sidewall}(d), left).

A mirror-symmetric $1{\times}2$ splitter provides the control. Across three
stack-specific fixtures and five angles, split balance remains between
$0.4982$ and $0.5072$, while SOI transmission moves from $0.787$ at
vertical walls to $0.935$ at $15^{\circ}$ and $0.581$ at $45^{\circ}$
(Fig.~\ref{fig:sidewall}(d), right). Sidewall tilt therefore changes an
asymmetry-dependent coupling ratio while preserving a symmetry-protected
balance. Four low-angle SOI fixtures have monitored output sums up to
$1.084$ at the draft grid. Normalization by total monitored output reduces
sensitivity to a common multiplicative error. The resolved quantities are
the angular trends and power ratios.

Each physical study retains its program, stack and specification files,
executed GDS, simulation inputs, cloud task identifiers, and evaluator
outputs. The representation bounds reachable geometry, while the evaluator
and modeled fabrication variables bound the physical conclusion.

\section{Dataset creation and training with verifier-derived signals}
\label{sec:rl}

\begin{figure*}[!t]
\centering
\includegraphics[width=\linewidth]{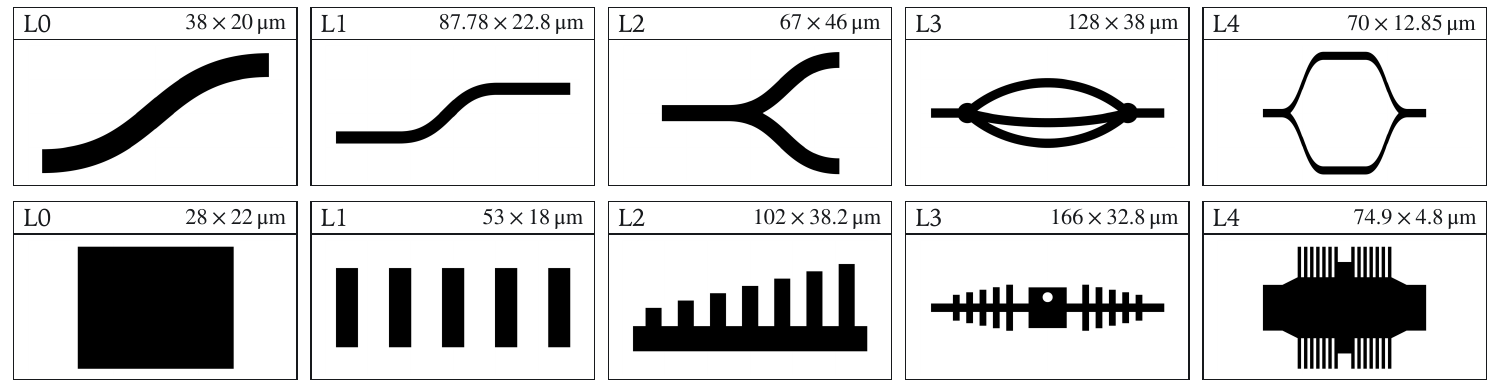}
\caption{\textbf{Representation curriculum in the frozen PixCell Dataset.}
Each column shows two model-input examples from one curriculum level. Across
both rows, the examples progress from individual primitives and operations to
local compositions, structured geometries, and complete components. The L4
panels show a complete MZI and a Bragg-cavity representation. The panels use
maximum-visibility rendering and are not shown at a common scale. The headers
list the physical footprint paired with each image.}
\label{fig:datasetcurriculum}
\end{figure*}

The PixCell process that takes a visual component to a program also provides
a direct method for creating synthetic training data. Programs in the PixCell
DSL can be run to produce target silhouettes at known physical footprints for
geometries expressible in the language. Each program-to-image execution creates
a data point that binds an image and physical footprint to the code that
produced its geometry. The code then serves as the target for the inverse
image-to-program task performed by PixCell. We use this forward process to
synthesize a dataset across the representation language.

\subsection{Dataset construction and release}

The core curriculum here progresses from a primitive vocabulary to complete
component geometries (Fig.~\ref{fig:datasetcurriculum}). Here, L0 contains 177
examples of the 22 permitted primitive geometries along with their constructor
modes, orientations, and scale. L1 then progresses to include 215 examples of
operations including placement, transforms, connections, paths, Boolean
operations, repetition, arrays, and routing. L2 contains 118 local compositions
such as connected chains, branches, repeated carriers, arrays, and radial
banks. L3 includes 120 structured examples with multiple zones, routes,
repeated media, defects, and cyclic organization. Finally, L4 contains 108
complete synthetic component geometries across couplers, MMIs, splitters,
interferometers, resonators, crossings, gratings, cavities, converters, and
free-propagation structures. Together, these five levels form a 738-row core
with 547 representation anchors and 191 supporting examples. In order to add
parameter depth without changing this grammar, the depth builder also
identifies live numeric dimensions in 546 of the 547 anchors and
deterministically produces seven additional settings for each representation.
The remaining anchor has no visible parameter variation under the rendering
policy. Five settings per expanded representation join the 738 core rows in
the training split, and two are held out for validation. The resulting depth
configuration contains 3{,}468 training and 1{,}092 validation examples. Its
validation split measures recovery at unseen parameter settings of known
representations rather than transfer to unseen representations. We release
this corpus as \texttt{qpaig-mit/pixcell} at the immutable \texttt{v2.0.0}
revision~\cite{pixcelldataset} with the \texttt{depth} configuration providing
4{,}560 model examples and \texttt{core} the 738-row curriculum subset. Each
model row contains the maximum-visibility image, physical footprint, program,
curriculum identifiers, and integrity hashes. A separate \texttt{references}
configuration stores the physical aspect target raster, calibration, ports,
topology, parameters, and lineage used by the evaluator and remains outside
model prompts. All images and programs in the release are synthetic but pass
structural checks, calibrated geometric comparison, and footprint agreement.

\subsection{Training with verifier-derived signals}

\begin{figure*}[!t]
\centering
\includegraphics[width=\linewidth]{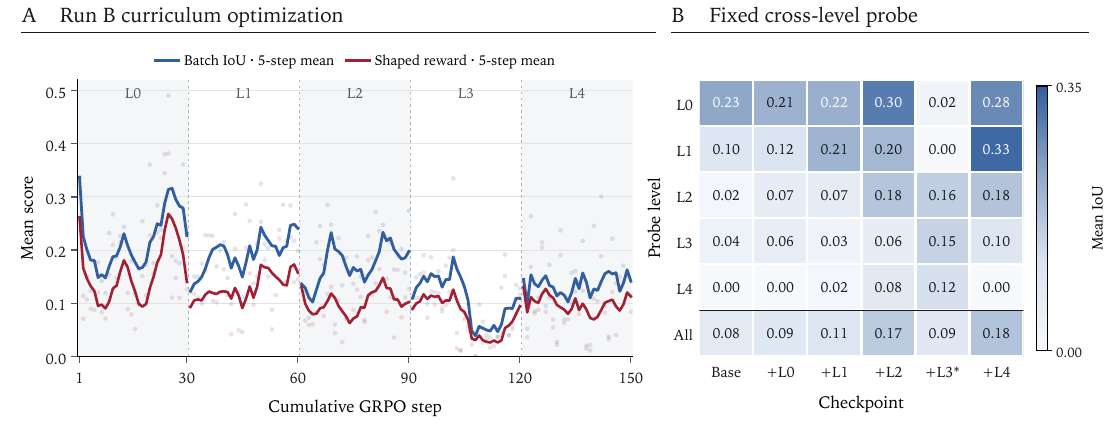}
\caption{\textbf{Run B training across the PixCell representation
curriculum.} (a) Batch IoU and shaped reward across 150 GRPO updates. Points
show individual updates and lines show five-step trailing means reset at each
level. The task distribution changes at every boundary, so discontinuities
also reflect the active curriculum level. (b) Mean IoU on the fixed 80-task
probe at the base checkpoint and after every stage. Each cell contains one
attempt on each of 16 held-out parameter settings at the indicated level. The
All row reports the mean across all 80 tasks. The archived post-L3 probe,
marked with an asterisk, was measured during a harvest and evaluation
contention window and is retained as recorded.}
\label{fig:rltraining}
\end{figure*}

The same deterministic evaluator used to verify reconstructed programs also
provides a reward for training. Run B begins from the raw
Qwen3.6-35B-A3B base checkpoint~\cite{qwen36}, applies a rank-32
low-rank adaptation (LoRA)~\cite{lora}, and completes 150 group relative
policy optimization (GRPO) steps~\cite{grpo} without supervised demonstrations
or critic-generated feedback. Each episode presents one
maximum-visibility image, its physical footprint, the permitted DSL catalog,
and the source contract. Following this, the model returns one program in a
non-thinking 4{,}096-token response, and the evaluator runs a process of
execution and comparison with the reference wherein syntax failures, source
violations, execution failures, and missing GDS outputs receive zero reward.
Moreover, reference or evaluator failures abort the run. The general motivation
behind such a direction is to have a signal that can cover the representation
space and determine whether small models can learn the representation. To that
extent, there is less focus on the manner in which signals for different types
of violations could be developed, as we focus on gauging learning capabilities.
To that end, for a source-compliant executable program, let $J$ and $D$ denote
IoU and Dice, $c$ the symmetric boundary chamfer distance in micrometers, and
$d$ the footprint diagonal. Here, the subscript $\mathrm{rect}$ denotes the
result from a solid rectangle filling the target's calibrated foreground
bounding box. Define
\begin{align*}
N(x;x_0) &= \max\left\{0,\frac{x-x_0}{1-x_0}\right\},\\
T(c) &= \exp\left[-\frac{c}{0.05d}\right],\\
B(c) &= \frac{\max\{0,T(c)-T(c_{\mathrm{rect}})\}}
{\max\{1-T(c_{\mathrm{rect}}),0.3\}}.
\end{align*}
\begin{subequations}\label{eq:rlreward}
\begin{samepage}
For $x_0=1$, $N(x;x_0)$ is defined as zero. When
$J_{\mathrm{rect}}<0.95$, the training reward is
\begin{align}
R ={}& 0.05 + 0.40N(J;J_{\mathrm{rect}}) \nonumber\\
&+ 0.15N(D;D_{\mathrm{rect}}) + 0.40B(c).
\label{eq:rlreward-general}
\end{align}
\end{samepage}
For rectangle-like targets with $J_{\mathrm{rect}}\geq0.95$, the reward is
\begin{equation}
R = 0.05 + 0.40J + 0.15D + 0.40T(c).
\label{eq:rlreward-rectangle}
\end{equation}
\end{subequations}
The rectangle baseline removes geometric credit for filling the target
bounding box, while the boundary term provides a signal for count and boundary
changes that may not yet improve overlap.

Training uses the Tinker API~\cite{tinker} throughout, with learning rate $10^{-5}$,
temperature $1.0$, eight task groups per step, and eight rollouts per group.
Within-group advantages are optimized with importance sampling and zero KL
penalty, and constant-reward groups are omitted. Run B spends 30 steps on each
level from L0 through L4 and initializes every stage from the preceding
checkpoint. After L0, each five-step cycle draws exactly 80\% of its task
groups from the active level and 20\% from earlier levels. The stage sequence
therefore contains 150 updates over 9{,}600 sampled programs.

The fixed probe uses the same 80 examples at the base checkpoint and after
each stage, with 16 rows from the validation split at every curriculum level
(Fig.~\ref{fig:rltraining}(b)). Between the base and final checkpoints, its
overall executable rate rises from $0.20$ to $0.55$, mean IoU from $0.079$ to
$0.179$, and mean shaped reward from $0.047$ to $0.124$. The final checkpoint
improves the base mean on L0 through L3, although none of its 16 L4 responses
executes on this probe draw. These rows hold out parameter settings of known
representations, so the probe measures recovery across the curriculum rather
than transfer to unseen representations.

We evaluate the final policy separately on the training-excluded F1--F8
targets used throughout the paper. The non-thinking base model produces no
executable program in 64 attempts. A direct draw from the trained policy
produces 39 executable programs in 64 attempts, with mean IoU $0.228$ and
mean best-of-eight IoU $0.467$ (Table~\ref{tab:rl}). The deployment loop uses
an independent draw of eight initial attempts per target, retains the
strongest candidate, and gives it three rounds of four revisions. Here, feedback
reports an error class, a closeness category, and whether material is missing
or excessive without exposing numerical measurements. Mean champion IoU
rises from $0.422$ after the initial attempts to $0.452$, $0.476$, and $0.491$
after the three revision rounds. This is one trained lineage evaluated
through repeated samples and revisions.

\begin{table}[t]
\caption{\textbf{Run B evaluation on the training-excluded F1--F8 targets.}
The single-pass draw contains eight independent attempts per target at
temperature $1.0$. Mean IoU includes failed programs as zero and best of 8
retains the highest. The iterative champion comes from a separate
eight-attempt draw followed by three rounds of four revisions under
number-free verifier feedback.}
\label{tab:rl}
\begingroup
\scriptsize
\setlength{\tabcolsep}{3.2pt}
\renewcommand{\arraystretch}{1.06}
\begin{ruledtabular}
\begin{tabular}{@{}lcccc@{}}
Device & Executable & Mean IoU & Best of 8 & Iterative \\
\colrule
F1 & 7/8 & 0.247 & 0.436 & 0.518 \\
F2 & 8/8 & 0.429 & 0.523 & 0.507 \\
F3 & 4/8 & 0.082 & 0.214 & 0.336 \\
F4 & 5/8 & 0.314 & 0.663 & 0.651 \\
F5 & 3/8 & 0.063 & 0.244 & 0.293 \\
F6 & 3/8 & 0.202 & 0.643 & 0.716 \\
F7 & 2/8 & 0.110 & 0.483 & 0.489 \\
F8 & 7/8 & 0.375 & 0.526 & 0.418 \\
\colrule
All / mean & 39/64 & 0.228 & 0.467 & 0.491 \\
\end{tabular}
\end{ruledtabular}
\endgroup
\end{table}

The released LoRA adapter is available as
\href{https://huggingface.co/qpaig-mit/pixcell}{\texttt{qpaig-mit/pixcell}}.
Its validated benchmark and revision records are released with the code.
Together, the fixed synthetic probe and the training-excluded paper targets
show that executable geometric rewards improve program generation across the
representation curriculum while leaving substantial room between the trained
policy and the coding agents of Sec.~\ref{sec:results}.

\section{Conclusion and research contracts}\label{sec:conclusion}

PixCell establishes a visual-to-executable interface for photonic component
creation where multimodal agents can successfully represent visual inputs as
parametric code composed of geometric primitives. The deterministic
verification metric makes those programs measurable, and their parametric
nature connects reconstruction to cross-stack evaluation and the possibility
of PDK retargeting. This interface also enables synthetic dataset creation and
verifier-derived training. As a result, we have connected representation,
verification, retargeting, simulation, and training through one program
interface. The results leave four research directions that matter directly to
what PixCell can establish. We release them as research contracts, with the
versioned contract files defining the current tests, evidence requirements,
and claim boundaries:
\begin{enumerate}
\item \textbf{RC-01, shape-aware visual verification,} extends geometric scoring so
structurally correct reconstructions can be separated from visually similar
shortcuts. Verification guides reconstruction, selection, and training
throughout PixCell, and so better verification metrics are crucial.
\item \textbf{RC-02, process-faithful 2D-to-3D retargeting,} carries editable programs
into fuller stack and fabrication models. This connects the parametric freedom
demonstrated in Sec.~\ref{sec:pdk} to physical conclusions beyond
simplified two-dimensional geometry.
\item \textbf{RC-03, representation and scale dataset,} expands the data linking
topology, physical-scale evidence, and executable construction. This supports
controlled study of what a model learns from pixels, calibration, and program
structure.
\item \textbf{RC-04, smallest qualifying open model,} maps how model scale
affects reconstruction across the representation curriculum. This extends the
single open-weight training lineage studied here towards smaller,
reproducible systems.
\end{enumerate}

The current versioned terms, evidence requirements, and executable verdict
logic for these directions are released under
\href{https://github.com/QPG-MIT/PixCell/tree/main/research-contracts}
{\texttt{research-contracts/}} in the PixCell repository. The
\href{https://pixcell.qpaig.com/\#research-contracts}{PixCell webpage}
shows live funding and claim state, links each specification, and provides the
claim controls. Importantly, our recomputation of the required evidence
determines settlement of these contracts.

\paragraph*{Data and code availability.} The PixCell implementation,
benchmark fixtures, 208 reconstruction programs, conductor measurements,
worker transcripts, retargeting records, simulation inputs and result
matrices, training recipes, and figure-generation scripts are released at
\url{https://github.com/QPG-MIT/PixCell}. The repository maps claims to
artifacts in \texttt{data/README.md}. Large FDTD fields and render rasters
are indexed by a checksummed external-bundle manifest. The frozen synthetic
curriculum is released at
\url{https://huggingface.co/datasets/qpaig-mit/pixcell} under revision
\texttt{v2.0.0}. Derived training tables and figure assets are included.

\begin{acknowledgments}
Artificial Intelligence, particularly Claude Opus 4.8, Fable 5, and ChatGPT
Codex 5.5 and 5.6 Sol, was substantively used in the drafting of this paper,
particularly in building the entirety of the codebase and contributing major
sections of the content. However, none of the ideas and motivations themselves
were proposed or directed by these models. The authors claim full
responsibility for the research and all associated claims and data presented
in this paper, all of which were created or certified by them.
This work was supported in part by the NSF National Quantum Virtual Laboratory
(NQVL:QSTD) under Award No.~\mbox{2533041} (ORAQL: Open-Stack Rydberg Atom Quantum
Computing Laboratory) and by the NSF Center for Quantum Networks under Award
No.~\mbox{EEC-1941583}.
We thank the Thinking Machines Lab team for supporting the model-training
experiments with a research grant and access to the Tinker API. We also thank
the Google DeepMind team for graciously extending the Gemini API rate limits
used in the reconstruction experiments.
\end{acknowledgments}

\bibliographystyle{apsrev4-2}
\bibliography{emergence}

\end{document}